\documentclass{article}

\usepackage[dblblindworkshop,final]{neurips_2026}

\usepackage[T1]{fontenc}
\usepackage{booktabs}
\usepackage{amsfonts}
\usepackage{amsmath}
\usepackage{amssymb}
\usepackage{microtype}
\usepackage{xcolor}
\usepackage{graphicx}
\usepackage{url}

\definecolor{linkcol}{HTML}{1F6E9C}
\usepackage[colorlinks=true,linkcolor=linkcol,citecolor=linkcol,
            urlcolor=linkcol,breaklinks=true]{hyperref}

\newif\ifshowtodo\showtodotrue
\newcommand{\authornote}[1]{\ifshowtodo\textcolor{red}{\textbf{[author: #1]}}\fi}

\newcommand{\ndcg}{nDCG@10}
\newcommand{\bpv}{B/vector}
\newcommand{\rates}{\mathcal{B}}
\newcommand{\dims}{\mathcal{M}}

\title{MRVQ: One Resident Index for Dimension- and Rate-Elastic Vector Search}

\workshoptitle{NeurIPS 2026 Workshop on Machine Learning for Systems (MLForSys)}

\author{%
  Sean Culatana \\
  Atlassian \\
  Mountain View, CA \\
  \texttt{sculatana@atlassian.com} \\
  \and
  \textbf{Shang-En Huang} \\
  National Taiwan University \\
  Taipei, Taiwan \\
  \texttt{sehuang@csie.ntu.edu.tw} \\
  \and 
  \textbf{Kang Li} \\
  Atlassian \\
  Bellevue, WA \\
  \texttt{kli3@atlassian.com}
}

\begin{document}

\maketitle

\begin{abstract}
Dense-retrieval services must switch among embedding-prefix dimensions and index bit rates as latency, quality, and
memory budgets change. Tuning a quantizer separately for each rate gives the best quality, but the retrieval tier then
holds several code streams and quantizer states at once. We introduce \textbf{Matryoshka Residual Vector Quantization
(MRVQ)}, a post-hoc residual quantizer for frozen embeddings. Its maximum-rate code can be truncated two ways:
dropping residual stages lowers the rate, and dropping embedding coordinates lowers the dimension. One resident
artifact therefore serves every (dimension, rate) pair we evaluate. Across FiQA and NFCorpus, four embedding families,
and $\{4,8,16\}$-byte codes, MRVQ is the lowest-RAM design we evaluate. It uses $17.8$--$22.0\times$ less memory than
three separately trained QINCo2 indices, and $1.89$--$2.02\times$ less than a lean shared-model steelman. The saving
is not free: per-rate QINCo2 is $0.026$--$0.107$~\ndcg{} better on FiQA. But MRVQ beats PQ, OPQ, and AdANNS-OPQ at
matched code size. We also evaluate a low-build-cost PCA-scalar design that attains quality comparable to RaBitQ and
its extension while fitting $420\times$ faster at the median. Finally, we report two negative results: QINCo2 collapses
when trained at high rates, and a ranking-bound hypothesis misses its pre-specified acceptance criteria. MRVQ is
therefore a low-memory operating point for elastic retrieval, not a universal quality winner.
\end{abstract}

\section{Introduction}
Dense retrieval is a first-class component of retrieval-augmented generation and agentic systems. For large corpora,
the document embeddings and index state---rather than the query encoder---often dominate resident memory: a
$768$--$1024$ dimensional fp32 vector occupies $3$--$4$~KB before indexing. That state must also support more than one
operating point. \emph{Dimension elasticity} uses prefixes of Matryoshka representations to trade retrieval quality for
query compute and bandwidth \citep{kusupati2022matryoshka,rege2023adanns}; \emph{rate elasticity} changes the document
code size to trade quality for storage. A high-quality deployment can tune a quantizer independently for each rate, but
then the retrieval tier retains several code streams and model states at once.

\begin{figure}[t]
\centering
\includegraphics[width=\textwidth]{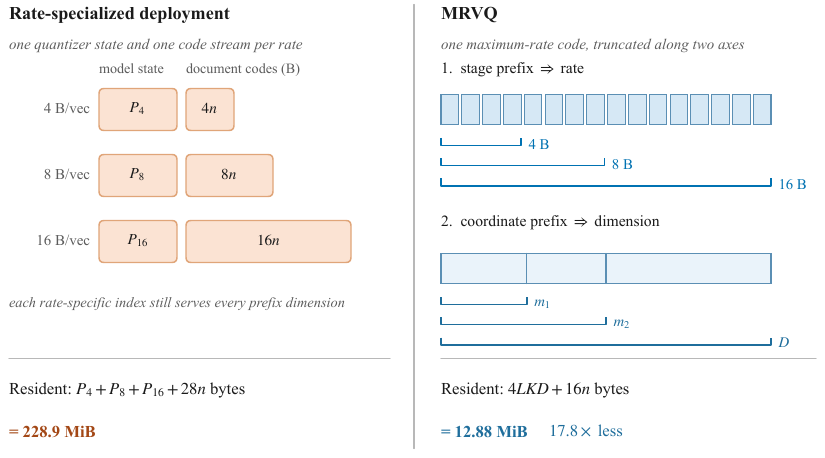}
\caption{\textbf{Amortizing a rate-and-dimension menu.} A rate-specialized deployment keeps one quantizer state and one
code stream per rate; each stream may still expose every embedding-prefix dimension. MRVQ stores one maximum-rate
residual code: stage prefixes select $4/8/16$~\bpv{}, and coordinate prefixes select the embedding dimension. The
document-code term falls from $28n$ to $16n$ bytes, and sharing the quantizer state produces the larger measured saving
(Table~\ref{tab:ram}). Quoted totals are FiQA at $D\!=\!768$.}
\label{fig:schematic}
\end{figure}

We ask whether a single resident document code can cover both axes. \textbf{MRVQ} fits a residual vector quantizer
post-hoc on frozen embeddings, weighting reconstruction error across the served prefix dimensions. Residual stages
provide nested rates, and coordinate prefixes provide nested dimensions, so the same stored code supports every
$(m,b)\in\dims\times\rates$ (Figure~\ref{fig:schematic}). This differs from AdANNS, which targets dimension-adaptive
search, and from MatQuant, which nests model-weight precision rather than document-index codes
\citep{nayak2025matquant}; Appendix~\ref{app:related} places MRVQ among quantizers for approximate search.

Our contribution is a systems trade-off, not a claim of universal quality dominance. First, we formalize
resident-memory accounting for a full rate menu and show that one MRVQ artifact is the lowest-RAM configuration
evaluated. Second, we measure the corresponding quality cost against QINCo2 and the gains over standard ANN
quantizers. Third, we evaluate a separate, inexpensive PCA-scalar design for hosts on which iterative quantizer
training is undesirable. Finally, we report two negative results---high-rate QINCo2 training failures and a
ranking-preservation predictor that does not meet our pre-specified bar.

\section{MRVQ and Resident-Memory Model}
\label{sec:method}
Let $\{x_i\}_{i=1}^{n}\subset\mathbb{R}^{D}$ be frozen document embeddings. An $L$-stage residual vector quantizer
with $K=256$ codewords per stage produces indices $z_i^{(1:L)}$ and reconstructions
$\hat{x}_i^{(\ell)}=\sum_{j=1}^{\ell}C^{(j)}_{z_i^{(j)}}$. Because each stage index occupies one byte, decoding the
first $b$ stages gives a $b$-byte code. MRVQ makes the same code dimension-elastic by fitting the codebooks under
\begin{equation}
\mathcal{L}_{\mathrm{prefix}}
=\sum_{i=1}^{n}\sum_{m\in\dims}w_m
\left\|x_i^{[:m]}-\hat{x}_i^{(L)[:m]}\right\|_2^2,
\qquad w_m\ge 0,\quad \sum_m w_m=1 .
\label{eq:prefix}
\end{equation}
At serving time, $(m,b)$ selects $\hat{x}_i^{(b)[:m]}$; no re-encoding and no additional document code is required.
Queries remain in floating point and are scored asymmetrically, as in PQ/OPQ and RaBitQ.

For the evaluated $L=16$, MRVQ's resident state is $R_{\mathrm{MRVQ}}=4LKD+16n$ bytes: fp32 codebooks plus one
$16$-byte code per vector. A deployment with a separate QINCo2 artifact for each $b\in\rates=\{4,8,16\}$ costs
$R_{\mathrm{sep}}=\sum_{b\in\rates}(P_b+nb)$, where $P_b$ is the corresponding model state; a shared-model steelman
costs $R_{\mathrm{shared}}=P_{16}+16n$. The document-code term alone therefore changes from $28n$ to $16n$ bytes
($1.75\times$ asymptotically), and at the corpus sizes we evaluate, sharing the fixed quantizer state produces a much
larger ratio. Appendix~\ref{app:scaling} reports the fixed and per-vector terms separately so the accounting can be
checked at any $n$.

We also study a separate \textbf{PCA-scalar} operating point: rotate by PCA, allocate scalar bits by coordinate
variance, and quantize each coordinate independently. It requires an eigendecomposition but no iterative
learned-codebook training; it is not included in the MRVQ memory accounting.

\section{Evaluation}
\label{sec:setup}
We use FiQA ($57{,}638$ documents, $648$ queries) and NFCorpus ($3{,}633$ documents, $323$ queries) from BEIR
\citep{thakur2021beir}. The four embedding families are MPNet \citep{reimers2019sbert} ($D=768$), Mxbai
\citep{lee2024mxbai} ($D=1024$), Nomic \citep{nussbaum2024nomic} ($D=768$), and BGE \citep{xiao2023bge} ($D=768$).
Within each family, every method uses the same fixed, model-specific prefix menu $\dims\subseteq
\{128,256,384,512,768,1024\}$, and a ``menu mean'' is the unweighted mean over that family's evaluated prefix-rate
cells. The prefix weights $w_m$ in Eq.~\eqref{eq:prefix} are uniform over $\dims$ and all $16$ residual stages are
weighted equally; Appendix~\ref{app:config} lists the exact checkpoints, menus, and fitting budget. Rates are
$\rates=\{4,8,16\}$~\bpv{}, or $192$--$1024\times$ smaller than the corresponding fp32 vector before fixed quantizer
overhead.

The metric is \ndcg{}. Learned baselines use matched fitting budgets, and their aggregated per-rate results reproduce
the stored runs to floating-point tolerance (maximum absolute deviation $0$ for MRVQ, $5\times10^{-17}$ for QINCo2). We
use paired per-query bootstrap confidence intervals ($5000$ resamples; $10{,}000$ for the RaBitQ study). Decision
thresholds for the negative-result tests were fixed before the final aggregate analyses. Reported RAM is analytical
accounting from the byte formulas above---parameter and code bytes only, with no optimizer or runtime buffers---rather
than measured process RSS.

\section{Results}
\label{sec:results}
\begin{figure}[t]
\centering
\includegraphics[width=\textwidth]{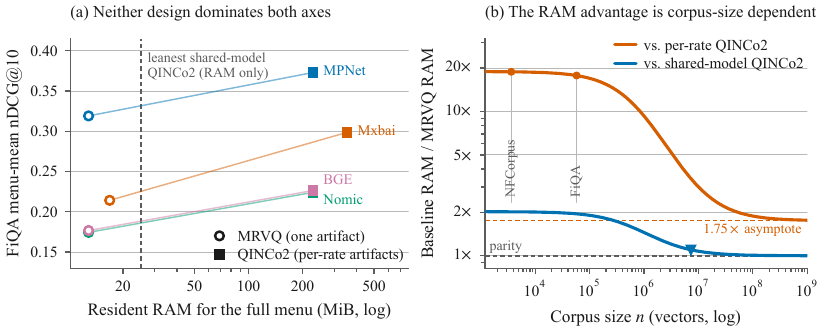}
\caption{\textbf{(a)} FiQA menu-mean \ndcg{} versus resident RAM for the full menu. MRVQ ($\circ$) occupies the
low-memory corner and per-rate QINCo2 ($\square$) the high-quality corner; neither dominates on both axes. The dashed
line marks the leanest shared-model QINCo2 RAM, whose quality we do not measure. \textbf{(b)} The same accounting as a
function of corpus size at $D\!=\!768$. The advantage over per-rate QINCo2 decays from $\sim$$18\times$ at the
evaluated corpora to the $1.75\times$ code-stream asymptote, and the shared-model advantage falls below $1.1\times$ at
$n\!=\!7.3$M ($\blacktriangledown$).}
\label{fig:pareto}
\end{figure}

\paragraph{One artifact is the low-RAM corner.}
Table~\ref{tab:ram} prices the complete $\{4,8,16\}$~\bpv{} menu. MRVQ uses $12.06$--$16.88$~MiB and is the lowest-RAM
configuration for all four families on both datasets. Relative to three separate QINCo2 models, the saving is
$17.8$--$22.0\times$; relative to the leanest shared-model configuration (including reduced embedding width and fp16
variants), it is $1.89$--$2.02\times$. The first ratio is largest at these on-device and per-tenant corpus scales
because neural parameter state dominates the resident footprint (Appendix~\ref{app:scaling}). As $n$ grows, the
separate-model ratio approaches the $28/16=1.75\times$ code-stream ratio, while the shared-model gap becomes a fixed
overhead and therefore less important; Figure~\ref{fig:pareto}(b) makes both limits explicit.

\begin{table}[t]
\centering
\small
\caption{Resident RAM (MiB) for the complete $\{4,8,16\}$~\bpv{} menu; ratios are relative to MRVQ. ``Separate''
retains one QINCo2 model and code stream per rate; ``shared'' is the leanest single-model steelman evaluated. Families
that share an embedding width share a row because the accounting depends on $D$, not on the family.}
\label{tab:ram}
\begin{tabular}{llrrrr}
\toprule
Dataset & Family & $D$ & MRVQ & QINCo2 separate & QINCo2 shared \\
\midrule
FiQA     & MPNet/Nomic/BGE & 768  & 12.88 & 228.9 ($17.8\times$) & 25.17 ($1.95\times$) \\
FiQA     & Mxbai           & 1024 & 16.88 & 354.6 ($21.0\times$) & 31.85 ($1.89\times$) \\
\addlinespace[2pt]
NFCorpus & MPNet/Nomic/BGE & 768  & 12.06 & 227.4 ($18.9\times$) & 24.34 ($2.02\times$) \\
NFCorpus & Mxbai           & 1024 & 16.06 & 353.2 ($22.0\times$) & 31.03 ($1.93\times$) \\
\bottomrule
\end{tabular}
\end{table}

\paragraph{Quality forms a two-corner Pareto.}
At matched code size, separately trained QINCo2 \citep{huijben2024qinco,vallaeys2025qinco2} beats MRVQ in every FiQA
family-rate cell by $0.026$--$0.107$~\ndcg{} (Appendix~\ref{app:quality}). Conversely, even the shared-model QINCo2
steelman uses roughly $1.9\times$ MRVQ's RAM at the evaluated corpus sizes. Figure~\ref{fig:pareto}(a) therefore shows
two useful operating points rather than a single winner. The low-memory result repeats on NFCorpus for all four
families.

Against standard fixed-code ANN baselines, MRVQ is stronger at matched bytes. Across the aggressive $4$--$16$~\bpv{}
regime, paired-bootstrap intervals favor MRVQ over per-prefix OPQ by about $0.06$--$0.07$~\ndcg{}, over PQ by
$0.10$--$0.12$, and over AdANNS-OPQ \citep{rege2023adanns} in every evaluated cell. The practical rule is therefore
constraint-dependent: choose QINCo2 when quality dominates, and MRVQ when resident memory prevents keeping several
specialized artifacts.

\paragraph{A low-build-cost scalar alternative.}
We compare PCA-scalar with RaBitQ and Extended-RaBitQ \citep{gao2024rabitq,gao2024extendedrabitq} using native
Faiss~1.13.2 asymmetric cosine search on $\ell_2$-normalized vectors. At nominal $b\in\{1,2,4\}$ bits/dimension, the
RaBitQ representation receives an additional $8/20/20$~\bpv{}, so the byte accounting is conservative for PCA-scalar.
Across the two aggregate query grids, pooled PCA-scalar $-$ RaBitQ deltas are $+0.0076/+0.0029/+0.0010$ and
$-0.0009/+0.0012/+0.0006$~\ndcg{} at $b=1/2/4$. All absolute effects are below $0.008$, and one Mxbai $1$-bit cell
favors RaBitQ by $0.0039$; we therefore describe quality as comparable, not superior. Timed separately on three
held-out BEIR corpora ($18$ cells, fixed $8$-thread CPU), PCA training is $420\times$ faster than matched OPQ fitting
at the median and $686\times$ at the mean, with a $47$--$1967\times$ spread: $0.17$--$6.95$~s against
$72$--$416$~s. That makes it attractive for in-place rebuilds (Appendix~\ref{app:rabitq}).

\section{Limitations and Conclusion}
\label{sec:limits}
\paragraph{Training reliability.} On NFCorpus at $16$ stages, per-rate QINCo2 suffers retrieval collapse---\ndcg{}
below half the MRVQ value despite converged validation MSE---in $3/3$ BGE seeds (best $0.035$) and $1/3$ Nomic seeds.
The best surviving Nomic seed is also non-monotone in rate: $0.294$ at $16$ stages against $0.320$ at $8$. We retain
these failures rather than silently dropping them.

\paragraph{No ranking certificate.} A parameter-free predictor based on residual energy is directionally useful but
misses our pre-specified acceptance bar: pooled rank correlation is $0.825$ against a required $0.85$, and only $72\%$
of cells yield a non-vacuous bound against a required $90\%$ (Appendix~\ref{app:p4}).

\paragraph{Scope.} The quality study covers two modest BEIR corpora and four embedding families. We measure resident
memory and ranking quality, not end-to-end ANN latency, build energy, or recall--throughput curves; the RAM gains
should not be read as speedups. At very large $n$ the advantage over a shared-model baseline narrows as the accounting
predicts.

Within that scope, MRVQ converts a dimension-and-rate menu from several rate-specialized resident artifacts into one
post-hoc code. It pays a measurable quality cost to QINCo2, but beats standard ANN quantizers at matched bytes and
occupies a distinct low-memory Pareto point. That is the intended systems contribution: a simple deployment option for
elastic retrieval when resident memory, rather than peak quality, is the binding resource.

\bibliographystyle{plainnat}
\bibliography{refs}

\appendix
\section{Resident-memory accounting and scaling}
\label{app:scaling}
\begin{figure}[h]
\centering
\includegraphics[width=\textwidth]{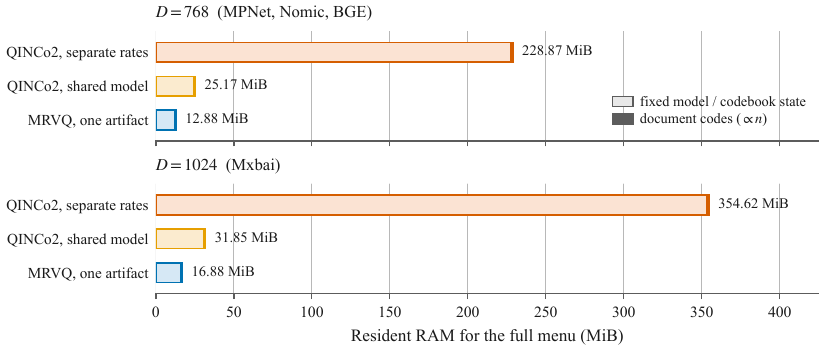}
\caption{FiQA ($n=57.6$k) resident RAM for the full rate menu, split into fixed model or codebook state and the
document codes that grow with $n$. The code term is a thin sliver at this corpus size, which is why the measured ratios
in Table~\ref{tab:ram} are far above the asymptotic $1.75\times$.}
\label{fig:breakdown}
\end{figure}

Table~\ref{tab:accounting} separates each configuration's resident state into a fixed term and a per-vector term, so
Table~\ref{tab:ram} can be reconstructed at any corpus size. The MRVQ fixed term is $4LKD$ bytes exactly. The
separate-rate QINCo2 term counts three rate-specific models ($d_e\!=\!D$, $d_h\!=\!256$, $K\!=\!256$, $A\!=\!16$,
$B\!=\!32$, two sub-layers, fp32); at $D\!=\!768$ these hold $7.28$M, $16.72$M, and $35.60$M parameters, so
$4\sum_b P_b = 227.3$~MiB. The shared-model steelman is a single maximum-rate QINCo2 with $d_e\!=\!128$ in fp16,
prefix-truncated to serve $4/8/16$~\bpv{}. All terms count parameter bytes only. Figure~\ref{fig:breakdown} shows the
resulting split at FiQA scale, where the fixed state accounts for over $93\%$ of resident bytes in every
configuration.

Because MRVQ and the shared-model steelman carry the same $16n$ slope, their difference is a constant $12.29$~MiB at
$D=768$: MRVQ stays more than $1.1\times$ smaller through $n\approx7.3$M vectors, and the gap is under $8\%$ beyond
about $10$M vectors. The separate-rate menu instead carries a $28n$ slope, so its ratio to MRVQ decays toward
$28/16=1.75\times$ rather than to parity.

\begin{table}[h]
\centering
\small
\caption{Resident-state decomposition behind Table~\ref{tab:ram}. $L=16$ stages, $K=256$ codewords per stage.}
\label{tab:accounting}
\begin{tabular}{llrrr}
\toprule
 & & \multicolumn{2}{c}{Fixed term (MiB)} & Per-vector \\
\cmidrule(lr){3-4}
Configuration & Resident state & $D=768$ & $D=1024$ & (\bpv{}) \\
\midrule
MRVQ, one artifact     & $4LKD+16n$                  & 12.00  & 16.00  & 16 \\
QINCo2, separate rates & $\sum_{b\in\rates}P_b + 28n$ & 227.3  & 353.1  & 28 \\
QINCo2, shared model   & $P_{16}+16n$                & 24.29  & 30.97  & 16 \\
\bottomrule
\end{tabular}
\end{table}

\section{Full matched-byte quality grid}
\label{app:quality}
Table~\ref{tab:quality} reports FiQA MRVQ $-$ QINCo2 \ndcg{} at matched code size. All paired-bootstrap intervals favor
QINCo2; the values quantify the quality cost of sharing one code across the whole menu.

\begin{table}[h]
\centering
\small
\caption{FiQA MRVQ $-$ QINCo2 \ndcg{} at matched bytes. Negative values favor QINCo2.}
\label{tab:quality}
\begin{tabular}{lrrrr}
\toprule
Family & 4~\bpv{} & 8~\bpv{} & 16~\bpv{} & Rate mean \\
\midrule
MPNet & $-0.042$ & $-0.051$ & $-0.069$ & $-0.054$ \\
Mxbai & $-0.063$ & $-0.107$ & $-0.084$ & $-0.085$ \\
Nomic & $-0.042$ & $-0.057$ & $-0.050$ & $-0.050$ \\
BGE   & $-0.026$ & $-0.053$ & $-0.070$ & $-0.050$ \\
\bottomrule
\end{tabular}
\end{table}

\section{PCA-scalar versus RaBitQ}
\label{app:rabitq}
\begin{figure}[h]
\centering
\includegraphics[width=0.72\textwidth]{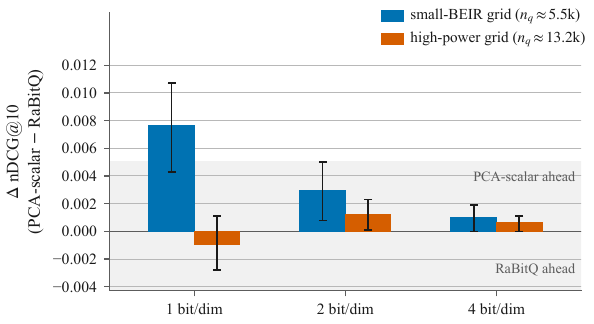}
\caption{PCA-scalar $-$ RaBitQ / Extended-RaBitQ \ndcg{} at matched nominal bits per dimension, with $95\%$
paired-bootstrap intervals. Positive values favor PCA-scalar; the shaded region is the pre-specified $\pm0.005$
practical-equivalence band.}
\label{fig:rabitq}
\end{figure}

Figure~\ref{fig:rabitq} plots the matched-nominal-bit deltas summarized in Section~\ref{sec:results}. The
pre-specified practical-equivalence band was $\pm0.005$~\ndcg{}. The pooled results are small, but the $1$-bit
small-grid mean and its whole confidence interval lie above that band, which is why the main text says ``comparable''
rather than claiming formal equivalence. Note also that the two grids disagree in sign at $1$ bit.

The fitting-speed comparison is timed on ArguAna, Quora, and SciDocs rather than on FiQA and NFCorpus, so it
characterizes the two fitting procedures rather than the specific indices evaluated above. Across those $18$ cells,
PCA training takes $0.17$--$6.95$~s against $72$--$416$~s for matched OPQ fitting; the ratio has median $420\times$
and mean $686\times$, ranging from $47\times$ to $1967\times$. We quote the median in the main text because the mean is
inflated by the largest corpus-model pairs.

\section{Ranking-preservation predictor}
\label{app:p4}
The tested predictor is $\epsilon(r)=\sqrt{D(r)}$, where $D(r)=\mathbb{E}_i\|x_i-x_i^{(r)}\|_2^2$ and $\|q\|_2=1$;
Cauchy--Schwarz then bounds RMS score perturbation by $\epsilon(r)$. The sweep covers $40$ cells: two datasets, four
families, and five rates $\{1,2,4,8,16\}$~\bpv{}. This rate set deliberately extends below the
$\rates=\{4,8,16\}$ serving menu of the main text, because a bound that is only meaningful at generous rates would be
of little use. Fixed-rate cross-geometry Spearman correlation is $0.66$, pooled predictor--gap Spearman correlation
is $0.825$, and only $72\%$ of cells have a non-vacuous bound. Both aggregate statistics miss the pre-specified
acceptance thresholds. Normalizing by $\mathbb{E}\|x\|_2^2$ removes a scale confound from unnormalized Mxbai
embeddings but does not change the verdict.

\section{Related work}
\label{app:related}
PQ, OPQ, and additive quantization are standard fixed-rate compressors
\citep{jegou2011product,ge2013opq,babenko2014additive}. QINCo/QINCo2 use implicit neural codebooks to improve
rate--distortion performance \citep{huijben2024qinco,vallaeys2025qinco2}. RaBitQ and its extension provide
scalar-quantization methods with error guarantees \citep{gao2024rabitq,gao2024extendedrabitq}. Matryoshka
representation learning and AdANNS expose dimension elasticity \citep{kusupati2022matryoshka,rege2023adanns}, while
MatQuant nests model-weight bit widths \citep{nayak2025matquant}. MRVQ instead targets the resident document index:
one post-hoc code is truncated along both the residual-stage and the embedding-coordinate axis.

\section{Experimental configuration}
\label{app:config}
Table~\ref{tab:config} collects the settings needed to reproduce the reported numbers.

\begin{table}[h]
\centering
\small
\caption{Experimental configuration.}
\label{tab:config}
\begin{tabular}{ll}
\toprule
Setting & Value \\
\midrule
Datasets                & BEIR FiQA ($57{,}638$ docs, $648$ queries), NFCorpus ($3{,}633$ docs, $323$ queries) \\
\addlinespace[2pt]
MPNet                   & \texttt{sentence-transformers/all-mpnet-base-v2}, $D=768$ \\
Mxbai                   & \texttt{mixedbread-ai/mxbai-embed-large-v1}, $D=1024$ \\
Nomic                   & \texttt{nomic-ai/nomic-embed-text-v1.5}, $D=768$ \\
BGE                     & \texttt{BAAI/bge-base-en-v1.5}, $D=768$ \\
\addlinespace[2pt]
Prefix menu $\dims$     & MPNet, BGE: $\{128,256,384,512,768\}$; Mxbai: $\{128,256,512,768,1024\}$; \\
                        & Nomic: $\{128,256,512,768\}$ \\
Prefix weights $w_m$    & uniform, $w_m = 1/|\dims|$; all $16$ residual stages weighted equally \\
Rate menu $\rates$      & $\{4,8,16\}$~\bpv{} \\
MRVQ quantizer          & $L=16$ stages, $K=256$ codewords per stage, fp32 codebooks \\
MRVQ fitting            & $50$ iterations, $50{,}000$ sampled vectors, seed $0$ \\
\addlinespace[2pt]
Retrieval               & asymmetric scoring, fp32 queries; Faiss~1.13.2 for the RaBitQ study \\
Metric                  & \ndcg{} \\
Uncertainty             & paired per-query bootstrap, $5000$ resamples ($10{,}000$ for RaBitQ) \\
RAM accounting          & analytical, from the byte formulas in Table~\ref{tab:accounting} \\
Build-time protocol     & CPU, fixed $8$ threads, sequential timing; $3$ seeds $\times$ $3$ bit budgets \\
                        & $\times$ $\{$ArguAna, Quora, SciDocs$\}$ $\times$ $\{$MPNet, Mxbai$\}$ $= 18$ cells \\
\addlinespace[2pt]
Artifact                & \authornote{anonymous repository link} \\
\bottomrule
\end{tabular}
\end{table}

\end{document}